\documentclass[journal]{IEEEtran}

\usepackage[left=0.625in,right=0.625in,top=0.75in,bottom=1in]{geometry}
\usepackage{ifpdf}
\usepackage{cite}
\usepackage{array}
\usepackage{dblfloatfix}
\usepackage{url}
\usepackage{amsmath}
\usepackage{ragged2e}
\usepackage{graphicx}
\usepackage{verbatim}
\usepackage{float}
\usepackage{lipsum}
\usepackage{multicol, blindtext}
\usepackage{caption}
\usepackage{subcaption}
\usepackage{xcolor}
\definecolor{mypink2}{RGB}{0, 0, 255}
\definecolor{green}{RGB}{0, 128, 0}
\usepackage{multirow}
\usepackage{array}
\usepackage{tabularx}
\usepackage[acronym]{glossaries}
\usepackage[ruled,vlined]{algorithm2e}
\usepackage{algorithmic}
\usepackage{fancyhdr} 
\usepackage[utf8]{inputenc}
\usepackage{tcolorbox}
\usepackage{xcolor}
\usepackage{url}
\usepackage{hyperref}
\usepackage{amssymb}

\begin{document}

\title{\fontsize{21pt}{21pt}\selectfont From Prompts to Protection: Large Language Model-assisted In-Context Learning for Smart Public Safety UAV
}
\author{ Yousef~Emami,~\IEEEmembership{Senior Member,~IEEE,}
        Hao~Zhou,~\IEEEmembership{Senior Member,~IEEE,}
        Miguel~Gutiérrez Gaitán,~\IEEEmembership{Senior Member,~IEEE,}
        Kai~Li,~\IEEEmembership{Senior Member,~IEEE,}
        Luis~Almeida,~\IEEEmembership{Senior Member,~IEEE}
        and~Zhu~Han,~\IEEEmembership{Fellow Member,~IEEE,}

\thanks{Manuscript received 1 June 2025; revised 17 November 2025; accepted 16 February 2026. 
This research was funded by the STINGRAY Open Seed Fund UC research project and ANID CPS-RTC grant CIA250016.} 
\thanks{This work is funded by national funds through FCT – Fundação para a Ciência e a Tecnologia, I.P., and, when eligible, co-funded by EU funds under project/support UID/50008/2025 – Instituto de Telecomunicações, with DOI identifier <https://doi.org/10.54499/UID/50008/2025}
\thanks{Yousef Emami is with the Real-Time and Embedded Computing Systems Research Centre (CISTER), 4200-135 Porto, Portugal   (email: emami@isep.ipp.pt)}

\thanks{K.~Li is with Real-Time and Embedded Computing Systems Research Centre (CISTER), Porto 4249-015, Portugal, and also with the Department of Electrical and Computer Engineering, Carnegie Mellon University, Pittsburgh, PA 15213, USA (email: kaili@ieee.org).}

\thanks{Miguel Gutiérrez Gaitán is with Pontificia Universidad Católica de Chile (email:miguel.gutierrez@uc.cl)}
\thanks{Hao Zhou is with the School of Computer Science, McGill University, Montreal, QC H3A 0E9, Canada. (Corresponding author, email: hzhou098@uottawa.ca)}
\thanks{Luis Almeida is with Instituto de Telecomunicações, Faculdade de Engenharia, Universidade do Porto, Rua Dr. Roberto Frias, 4200-465 Porto, Portugal (email:lda@fe.up.pt)}
\thanks{Zhu Han is with the Department of Electrical and Computer Engineering, University of Houston, Houston, TX 77004 USA (e-mail: hanzhu22@gmail.com).  }
\vspace{-20pt}}

\maketitle

\thispagestyle{fancy}            
\chead{This paper has been accepted by IEEE Wireless Communications Magazine. } 

\renewcommand{\headrulewidth}{1pt}      
\pagestyle{plain}

\begin{abstract}
 A public safety Uncrewed Aerial Vehicle (UAV) enhances situational awareness during emergency response. Its agility, mobility optimization, and ability to establish Line-of-Sight (LoS) communication make it increasingly important for managing emergencies such as disaster response, search and rescue, and wildfire monitoring. Although Deep Reinforcement Learning (DRL) has been used to optimize UAV navigation and control, its high training complexity, low sample efficiency, and the simulation-to-reality gap limit its practicality in public safety applications. Recent advances in Large Language Models (LLMs) present a promising alternative. With strong reasoning and generalization abilities, LLMs can adapt to new tasks through In-Context Learning (ICL), enabling task adaptation via natural language prompts and example-based guidance without retraining. Deploying LLMs at the network edge, rather than in the cloud, further reduces latency and preserves data privacy, making them suitable for real-time, mission-critical public safety UAVs. This paper proposes integrating LLM-assisted ICL with public safety UAVs to address key functions such as path planning and velocity control in emergency response. We present a case study on data collection scheduling, demonstrating that the LLM-assisted ICL framework can significantly reduce packet loss compared to conventional approaches while also mitigating potential jailbreaking vulnerabilities. Finally, we discuss LLM optimizers and outline future research directions. The ICL framework enables adaptive, context-aware decision-making for public safety UAVs, offering a lightweight and efficient solution to enhance UAV autonomy and responsiveness in emergencies.
\end{abstract}


\IEEEpeerreviewmaketitle
\section{Introduction}
Uncrewed Aerial Vehicles (UAVs) play a crucial role in sectors such as energy, environmental monitoring, parcel delivery, and public safety. They are especially transformative in public safety applications, including post-disaster assessment, search and rescue missions, wildfire monitoring, border surveillance, and crime scene reconstruction, due to their agility, optimized mobility, and ability to establish reliable Line-of-Sight (LoS) communications. Public safety UAVs enhance situational awareness for police, firefighters, and other first responders, thereby improving incident management and overall operational efficiency.\cite{8682048}.
\par
UAVs can use Machine Learning (ML) techniques to optimize navigation and control, especially during emergencies. For example, they apply Deep Reinforcement Learning (DRL) to optimize flight trajectories based on victims’ needs or to identify casualties and efficiently collect environmental data\cite{10246260}. However, DRL solutions require complex training and face challenges such as the simulation-to-reality gap and low sample efficiency, making them unsuitable for the urgent demands of emergency scenarios. Therefore, lightweight and adaptive solutions are needed to respond effectively to emergencies and improve the success rate of rescue operations.
\par
Recent advancements in artificial intelligence, particularly in Large Language Models (LLMs), are driving significant transformations in UAV communications \cite{javaid2024large}. With strong reasoning and generalization capabilities, LLMs enable advanced understanding, flexible adaptation, and real-time responsiveness in diverse UAV scenarios \cite{wu2025llm}. Integrating LLMs into UAV communications offers a promising path toward greater autonomy, equipping UAVs with enhanced decision-making abilities and enabling more effective responses during emergencies.
\par
The conventional approach to deploying LLMs relies on cloud-based infrastructures, which often introduces communication delays and raises privacy concerns – issues that are particularly problematic for time-sensitive and mission-critical UAV operations\cite{yan2025hybrid}. A recent trend is to shift LLM deployment to the network edge through edge server integration. This approach enables LLMs to directly serve the computational and operational needs of edge devices such as UAVs. Notably, LLMs can learn new tasks using In-Context Learning (ICL) by the UAV, allowing rapid adaptation without retraining. For example, ICL can support key UAV functions such as path planning, velocity control, data collection scheduling, and power management, thereby enhancing real-time decision-making in emergencies. \cite{yang2024plug}.
\par
ICL is inherently lightweight because it eliminates the need for conventional model retraining and complex optimization. Unlike DRL or supervised methods that require extensive gradient updates, ICL adapts to new environments purely through prompt design and contextual demonstrations, minimizing computational and energy costs. Its training-free nature allows real-time responsiveness to dynamic UAVs or network conditions without simulation cycles. Moreover, ICL is compatible with compact, instruction-tuned models that can be quantized and deployed efficiently on edge devices, enabling low-latency reasoning at the network edge. By leveraging natural-language task formulations instead of handcrafted reward functions, ICL offers a simplified, interpretable, and resource-efficient alternative to conventional learning pipelines.
\begin{figure*}[h!]
  \centering
  \includegraphics[scale=0.6]{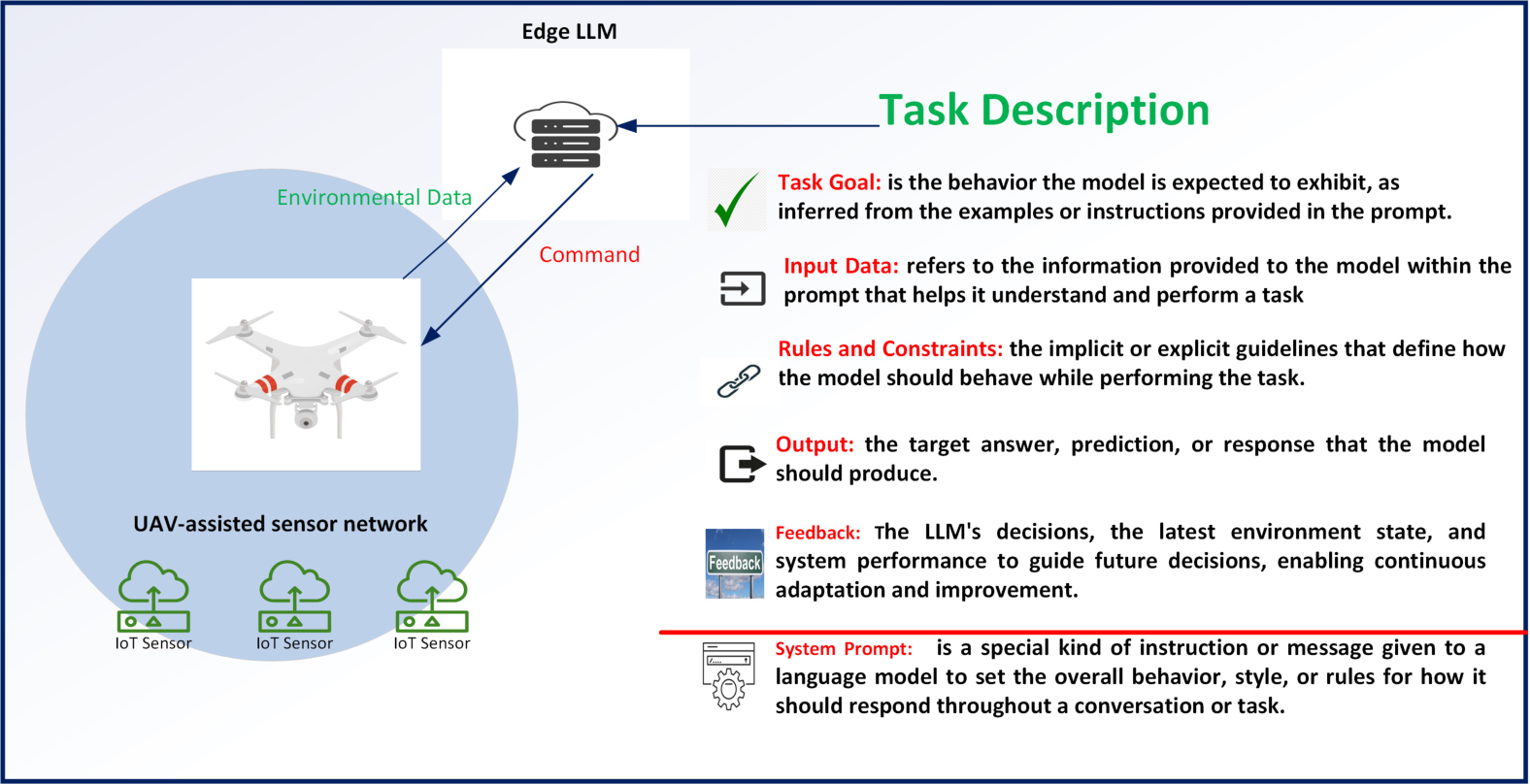}
  \caption{Illustration of LLM-assisted ICL for a public safety UAV. The public safety UAV collects sensory data and submits it to the edge LLM, where a task description, which is a basis for ICL, is formulated with the following components: task goal, input data, rules, output, and feedback. The system prompt defines the rule of LLM, e.g., data collector.}
  \label{fig:uav_network1}
\end{figure*}
\par

This paper advances the state of LLM-assisted UAV intelligence from conceptual discussions and single-task prototypes toward a fully deployable, network-aware control framework. Unlike prior studies that focus on descriptive overviews \cite{javaid2024large}, \cite{zhou2024large}, \cite{10835069} or isolated scheduling scenarios \cite{emami2025llm}, the proposed approach integrates in-network ICL with verifier-guided optimization to achieve safe, adaptive, and scalable multiple UAV functions. By coupling telemetry-driven prompt construction with a unified safety-verifier, the framework enables real-time context adaptation under dynamic wireless conditions while maintaining energy efficiency and decision robustness. Through quantitative evaluation across multiple performance metrics, this paper bridges the gap between reasoning-level intelligence and network-level optimization, demonstrating the practical feasibility of LLM-assisted UAV autonomy in real communication environments.As depicted in Fig. \ref{fig:uav_network1}, ICL leverages structured natural language prompts, i.e., task descriptions, to encapsulate task objectives, while curated demonstration sets convey relevant system states and environmental conditions such as sensory data. This approach enables LLMs to infer an effective output for the task, without requiring explicit model retraining. By using contextually rich examples, ICL supports adaptive, on-the-fly decision-making across a broad range of UAV-Assisted Sensor Network (UASNETs) tasks, making it particularly valuable for time-sensitive applications such as emergency response.

The contributions of this paper are summarized as follows: 

Firstly, we propose a unified framework and functional taxonomy for LLM-assisted ICL in public-safety UAVs. 
Unlike prior studies focusing on individual tasks such as data collection scheduling, this work establishes an integrated architecture that jointly supports key UAV functions such as path planning, velocity control, and power management. 
The framework provides a structured foundation and standardized blueprint for applying ICL in emergency contexts, particularly where DRL-based optimizers face high training complexity and limited real-world transferability. 
Secondly, we present a case study on data collection scheduling that demonstrates the effectiveness and security of the proposed ICL framework. 
Through a black-box attacker model, we analyze context-based jailbreaking threats and show that the ICL-based Data Collection Scheduling (ICLDC) system dynamically adapts to network variations while maintaining low packet loss and stable scheduling performance. 
This analysis highlights both the operational benefits and the resilience of LLM-assisted decision-making in mission-critical UAV environments. 
Thirdly, we provide a comparative discussion of optimization paradigms for public-safety UAVs, covering DRL, Diffusion Model (DM)+DRL hybrids, and LLM-assisted ICL. 
We further outline key research trends, including DM-based synthetic data generation, on-device LLM deployment, collaborative reasoning for UAV swarms, and real-world validation across domains. 
Together, these insights offer a practical roadmap for deploying LLM-assisted optimization in safety-critical UAV networks, balancing adaptability, efficiency, and robustness.

The remainder of this paper is organized as follows: Section \ref{sec2} discusses the deployment of LLMs. Section \ref{sec3} provides background on LLM-assisted ICL and its effectiveness. Section \ref{sec4} explores the application of LLM-assisted ICL for public safety UAV and its various functions. Section \ref{sec5} presents a case study on data collection schedule, along with investigating jailbreaking attacks. Section \ref{sec6} discusses LLM optimizers and their merits for public safety UAV. Section \ref{sec7} outlines future directions, including the use of DMs to enrich task descriptions. Finally, Section \ref{sec8} concludes the paper.

\section{Deployment of LLMs} \label{sec2}
This section highlights the importance of LLM deployment and discusses limitations of cloud-based deployment and the feasibility of edge-based LLM deployment. LLMs impose substantial demands on computational and storage resources. For example, GPT-4 contains approximately 1.76 trillion parameters and has a model size of 45 GB, placing significant strain on network storage infrastructure. Fine-tuning even a comparatively smaller LLM, such as a 7-billion-parameter model like GPT4-LLM, can require nearly three hours on a high-end setup with 8×80GB A100 GPUs—making the process highly time-intensive. Moreover, LLM inference contributes to overall network latency, influenced by factors such as hardware configuration, batch size, degree of parallelism, and optimization techniques like model pruning. Therefore, efficient and strategic deployment of LLMs is essential to meet the performance and scalability demands of UASNETs\cite{zhou2024large}.
\par
Thanks to their strong generalization and reasoning capabilities, LLMs enable UAVs to interpret complex, dynamic environments and develop coherent control strategies for navigation, surveillance, and decision-making tasks. These capabilities make LLMs well-suited for real-time operations in mission-critical scenarios such as disaster response, infrastructure inspection, and surveillance. However, most existing LLM deployments rely heavily on cloud-based infrastructures, which impose critical limitations in terms of latency, network bandwidth, and data privacy, making them suboptimal or even unfeasible for delay-sensitive or communication-limited UAV applications. To address this problem, efficient model compression techniques such as quantization, pruning, and knowledge distillation are used to minimize model size and computational load without compromising performance. In parallel, fast decoding strategies such as speculative decoding, early termination, parallel decoding, and sparse attention reduce inference latency and memory requirements. 
\begin{figure*} [h]
    \centering 
    \captionsetup{justification=raggedright}
    \includegraphics[width=18cm, height=8cm]{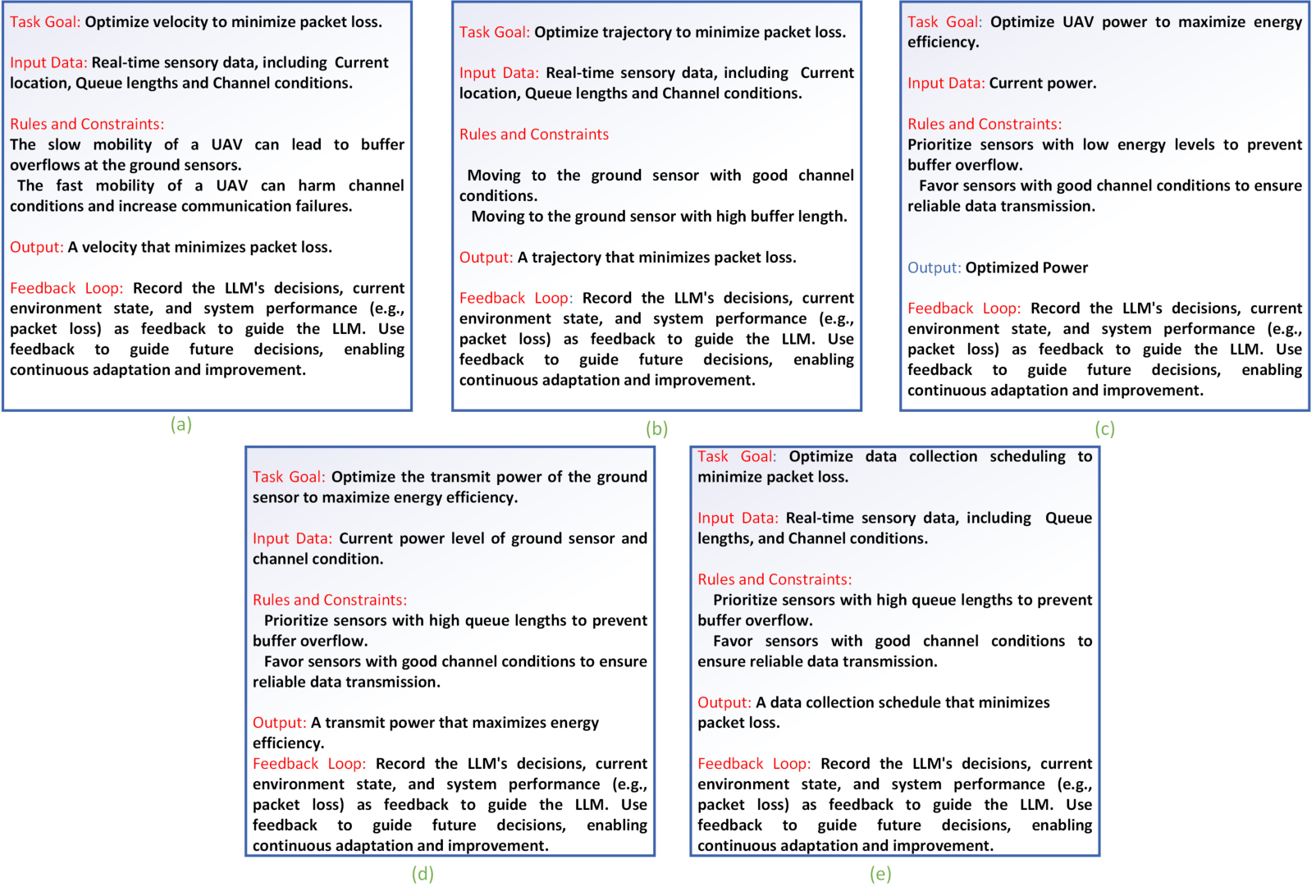}
    \caption{Task Descriptions for key functions in public safety UAV comprising task goal, input data, rules and constraints, output, and feedback loop.}
    \label{fig:task}
\end{figure*}

Once LLMs are efficiently deployed at the edge, they can be leveraged for context-aware inference through prompt engineering. Techniques such as CoT prompting, prompt-based planning, self-refinement prompting, and ICL enable LLMs to perform complex tasks by interpreting instructions and demonstrations embedded in prompts without the need for parameter updates. ICL is particularly effective in generalizing tasks because it uses previously trained knowledge and adapts to examples, with performance depending on the design of the prompts, including the selection, arrangement, and formatting of the demonstrations. In dynamic, resource-constrained scenarios such as UASNETs, ICL offers tangible benefits as previous solutions can serve as reusable inputs for new challenges, increasing the flexibility and responsiveness of UAVs\cite{10835069}.

\section{LLM-assisted In-Context Learning}  \label{sec3}
This section provides background on LLM-assisted ICL and discusses its effectiveness. The context used for ICL consists of an optional task instruction and a sequence of demonstration examples. 
The task instruction specifies the overall goal or nature of the task for the model. 
Each demonstration example consists of an input and its corresponding output, represented in natural language so that the model can interpret it. By conditioning on this context, the pretrained language model can perform a new task without updating its parameters.

\begin{figure}[h!]
  \centering
  \includegraphics[width=8cm, height=6cm]{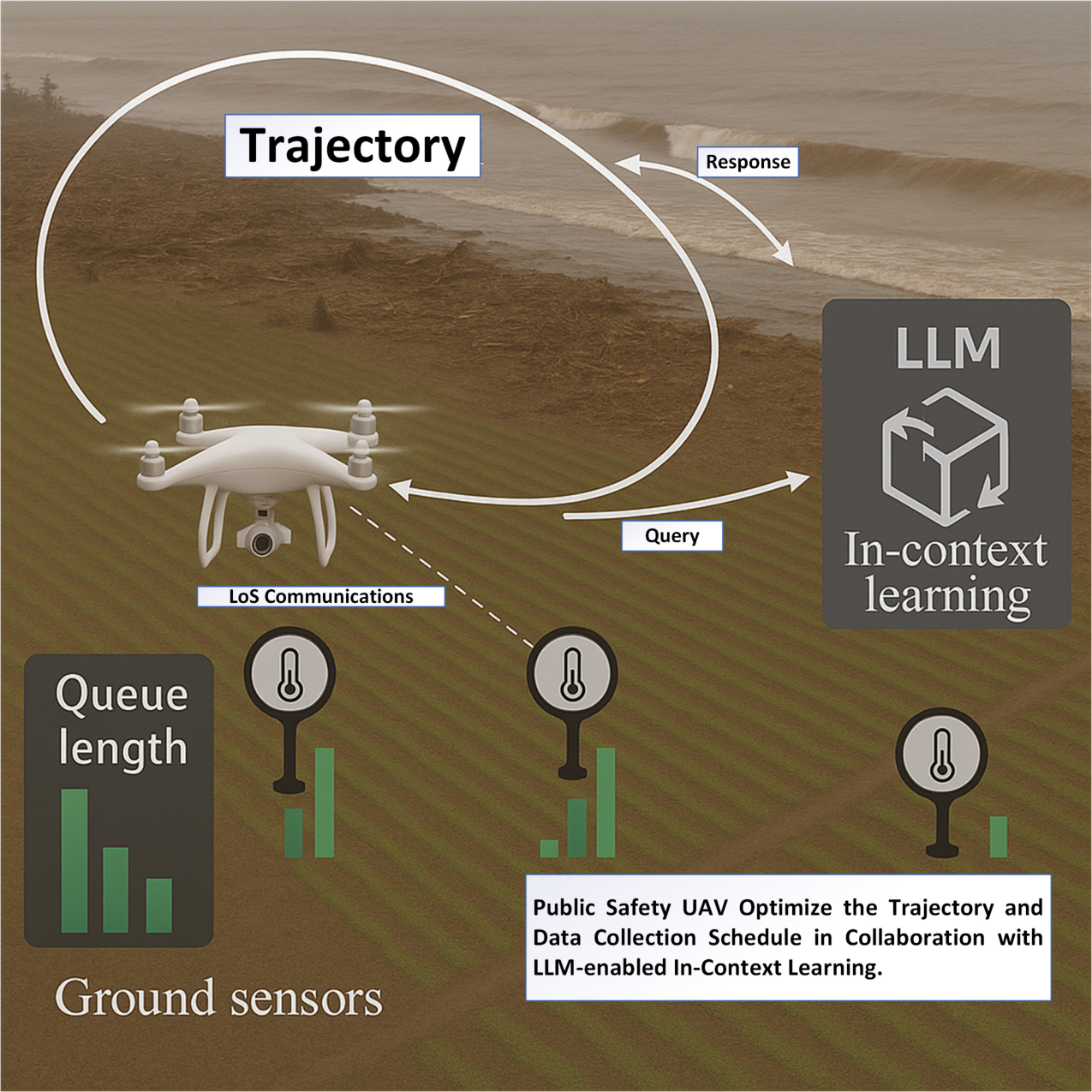} 
  \caption{Illustration of a public safety UAV, where the UAV follows the trajectory and establishes LoS communication with the ground sensor and uses the LLM as an optimizer.}
  \label{fig:uav_network}
\end{figure}

The examples in the context provide patterns for the model to follow, allowing it to generate an appropriate output for a new input query based on the information provided.

\begin{enumerate}
    \item \textbf{Prompt Learning:} While prompt learning uses fixed textual templates or soft prompt embeddings to elicit model behavior, ICL can be regarded as a subclass of prompt learning where the demonstrations are included directly in the task description. 
    
    \item \textbf{Few-Shot Learning:} Traditional few-shot learning involves adapting the model's parameters using a small number of supervised examples. In contrast, ICL requires no parameter updates and leverages the pretrained model directly through inference over the provided context\cite{dong2022survey}

\end{enumerate}

The likelihood of each candidate answer is determined using a scoring function, which evaluates how well the candidate aligns with the input query and the context provided by the task instruction and demonstration examples. Based on these scores, the model selects the candidate answer with the highest likelihood as its final prediction. In other words, the model compares all possible outputs in light of the context and chooses the one that best fits the pattern established by the examples and instructions, enabling accurate task completion without updating model parameters.
\par
Min et al.\cite{min2022rethinking} reveal that the effectiveness of ICL relies more on the structural presentation of input and label spaces than on the correctness of individual demonstration labels. Surprisingly, replacing ground truth labels with random ones in demonstrations has only a marginal effect on performance. Moreover, models can retain up to 95\% of performance gains using either inputs or label sets alone—provided the formatting is appropriate. These insights suggest that formatting and representational cues play a more critical role than previously assumed, with implications for both model design and future research in prompt engineering and meta-training for ICL.

\par
\section{LLM-assisted In-Context Learning for Optimized Public Safety UAV} \label{sec4}

This section highlights the potential of LLM-assisted ICL in optimizing key functions in public safety to dynamically adapt based on real-time context and prior experiences, all without the need for retraining the model, and provide discussion. We consider a post-disaster environment where an LLM is hosted at the network edge. The  UAV can physically approach individual ground sensors. The short, LoS-dominant communication link between a UAV and a ground sensor offers a significant channel gain and supports high-speed data transmission. In this context, employing a UAV for data collection enhances network throughput and extends coverage beyond what terrestrial gateways can achieve\cite{10278748}. 
\par
Throughout this section, the term ``feedback'' refers to externally measured performance indicators, such as packet loss, throughput, and energy consumption, which are computed by the UAV or the edge controller after each LLM-generated decision. These measurements are then summarized as contextual information and appended to the next prompt for further reasoning. The LLM itself does not calculate the numerical cost; instead, all costs are derived from observable system telemetry without relying on any ground-truth or oracle data.

Fig. \ref{fig:uav_network} illustrates a typical UASNETs, where ground sensors are deployed to monitor temperature and humidity in the disaster area. These ground sensors generate environmental data, which is temporarily stored in data queues for later transmission to the UAV. The UAV is deployed to hover over the disaster area and can maneuver close to each ground sensor, leveraging the short-distance LoS communication links to efficiently collect the stored data.

\subsection{Path Planning}

As depicted in Fig. \ref{fig:uav_network}, the UAV hovers over the disaster area where it adjusts its path relative to the requirements of the ground sensors. For example, the UAV must adjust its path based on channel conditions and the queue length of the ground sensors. Moving to the ground sensor with poor channel conditions or a small buffer length for data collection gives rise to packet reception errors or buffer overflow at other ground sensors. Therefore, online path planning of the UAV for data collection is crucial \cite{9148316}.  
\par
ML is essential for UAV placement and path planning due to the complex nature of designing UASNETs and the critical requirements in applications such as public safety. Determining the optimal horizontal and vertical placement of the UAV, as well as planning its trajectories to interact with ground or flying objects, requires DRL optimizers that can handle large-scale optimization problems and dynamic environmental factors. In a nutshell, DRL is indispensable for UAV placement and path planning and facilitates complex optimization tasks, supports real-time adaptation to changing conditions, and enhances capabilities for detecting and responding to noncompliant UAV in critical applications like public safety. Path planning involves the optimization of continuous variables. In this work, we simplify the problem by discretizing the action space and restricting the UAV's motion to a finite number of directions (e.g., north, south, east, and west). The use of a four-direction discrete action model prioritizes rapid, adaptive decision-making over aerodynamic precision, an essential trade-off in disaster-response missions. By constraining the action space to cardinal directions, the framework ensures computational tractability and real-time responsiveness for LLM-assisted reasoning, even on resource-limited UAV or edge platforms. This abstraction aligns with real-world UAV architectures, where high-level planners issue directional or waypoint commands while low-level controllers manage flight dynamics. Consequently, the discrete model serves as a practical and efficient bridge between natural-language reasoning and executable UAV control, enabling fast, mission-relevant adaptation in dynamic and safety-critical environments. 
Starting with a task description depicted in Fig. \ref{fig:task}(b), which defines the task and provides in-context demonstrations, the LLM as an optimizer generates candidate solutions. These solutions are then implemented in the environment, and their associated costs are computed. Feedback is derived from the latest sensory data, the implemented solution, and the obtained cost. This feedback, along with updated sensory data, is fed back into the LLM. The process repeats iteratively until satisfactory solutions are generated, enabling the LLM to refine its outputs dynamically based on real-world performance.

\par
In the path-planning task, the cost is defined as a combination of observable performance indicators, including the travel distance of the UAV, the measured packet-loss rate on the communication links, and the number of sensors whose data queues exceed a defined threshold. These values are obtained directly from UAV odometry, link-level acknowledgements, and queue reports from ground sensors, and then summarized in natural language as part of the feedback context for subsequent reasoning.

\subsection{Velocity Control}

As depicted in Fig. \ref{fig:uav_network}, the UAV hovers over the disaster area where it adjusts its velocity based on the requirements of the ground sensors. The slow mobility of a UAV can lead to buffer overflows at the ground sensors, as newly arriving data is not immediately captured by the UAV. The fast mobility of a UAV can harm channel conditions and increase communication failures. Onboard velocity control of the UAV is essential to minimize packet loss due to ground sensor buffer overflows and communication failures\cite{emami2023deep}. An LLM-assisted ICL framework can be employed to optimize the velocity of the UAV to minimize the overall packet loss. Starting from a task description depicted in Fig. \ref{fig:task}(a), which defines the velocity control task and provides in-context examples, the LLM acts as an optimizer to generate candidate velocities. The performance of the candidate velocities is evaluated by calculating a cost metric. The feedback is then shaped by the latest sensor data, the current velocity, and the calculated cost. This feedback is fed back to the LLM along with the current sensor data. The LLM iteratively refines its control strategies using this closed-loop process until the system converges to stable, optimal velocity tracking.
For velocity control, the cost metric emphasizes queue stability and communication reliability. It is derived from the observed frequency of queue overflows, the rate of transmission failures recorded through link acknowledgments, and a small penalty associated with abrupt changes in velocity. These values are collected by the UAV-edge controller and reported back as structured feedback, allowing the LLM to progressively refine its velocity adjustment policy.

\subsection{Data Collection Schedule}

As depicted in Fig. \ref{fig:uav_network}, the UAV hovers over the disaster area, enabling it to approach ground sensors closely and utilize short-range LoS communication links for efficient data collection. However, selecting a specific ground sensor for data collection may result in buffer overflows at other ground sensors if their data queues are near capacity while new data continues to accumulate. Moreover, ground sensors located farther from the UAV typically experience degraded channel conditions, increasing the likelihood of transmission errors. Therefore, an effective data collection schedule is essential to mitigate data queue overflows and reduce communication failures. LLM-assisted ICL is crucial for data collection scheduling in UASNETs, as it facilitates dynamic adaptation to changing environments. By leveraging ICL, the UAV can continuously adjust its data collection scheduling strategies in real- time, selecting the most appropriate sensors. This enables the system to make informed, context-aware decisions without retraining, thereby optimizing data collection efficiency and responsiveness in dynamic or mission-critical scenarios. This ensures efficient, reliable, and autonomous data collection operations, enhancing overall system performance. The relevant task description is depicted in \ref{fig:task}(e). This system uses an LLM to optimize data collection scheduling for minimal packet loss by processing real-time sensor inputs (queue lengths and channel conditions) while maintaining constraints that prioritize sensors with high queue lengths (to avoid buffer overflows) and those with good channel conditions (to ensure reliable transmission). The LLM generates an adaptive collection schedule, which is then deployed, recording the resulting system performance (packet loss) and the latest environmental state as feedback. This feedback loop, consisting of the LLM's decisions, the latest system state, and performance metric, enables continuous improvement by providing information for subsequent planning decisions, allowing the LLM to iteratively refine its policy to minimize packet loss.
In this scheduling task, the cost is evaluated using two observable components: the average packet-loss rate across all active sensors and the backlog penalty reflecting the number of sensors with nearly full data queues. Both quantities are directly measurable through MAC-layer acknowledgments and queue-level reports and are summarized by the edge controller as contextual feedback to guide the next scheduling decision.

\subsection{UAV Power Control}

We need to optimize energy efficiency and power control in UASNETs because the UAV and ground sensors have limited energy resources, making the optimization of energy and power consumption crucial for the overall performance of the UASNETs. DRL solutions have been proposed to exploit the network’s resources in a near-optimal manner, aiming to maximize energy efficiency effectively. However, these solutions require complex training. LLM-assisted ICL can be leveraged to optimize the transmit power and improve energy efficiency. 
\par
Task description for UAV power control is depicted in \ref{fig:task}(c) with a task objective, and the example set represents operational state and environmental context to assist the LLM in inferring optimal power control strategies. This system leverages an LLM to dynamically optimize UAV transmit power for maximum energy efficiency by processing real-time power levels and channel conditions, while adhering to constraints prioritizing sensors with low remaining energy (preventing node failure) and favoring those with strong channel conditions (ensuring reliable transmission). The LLM generates power adjustment commands executed, with the resulting system performance and updated operational state being captured as feedback. This closed-loop mechanism, incorporating the LLM's power decisions, the latest UAV status, and communication outcomes, enables continuous refinement of power allocation strategies, progressively allowing the UAV to achieve optimal energy efficiency. This approach enables adaptive decision-making without the need for complex training. Such optimization is particularly valuable in public safety scenarios such as emergency response, where efficient power usage directly impacts mission duration and system performance.
For UAV-side power control, the cost is computed as the ratio between the total energy consumed for transmission and the number of successfully delivered bits during a given observation window. Additional penalties are introduced when the observed packet-loss rate or latency exceeds predefined quality-of-service thresholds. These values are measured by the UAV’s onboard power monitor and link-layer counters and then summarized as text feedback to the LLM for subsequent optimization.

\subsection{Transmit Power of Ground Sensors}

Power control in UASNETs, which involves explicitly managing the transmit power of ground sensors, plays a key role in enhancing system performance. It optimizes interference management, signal quality, and resource usage, all while accounting for the limited battery capacities of the ground sensors. This system employs an LLM to optimize ground sensor transmit power for maximal energy efficiency. The LLM analyzes current power levels and channel conditions while adhering to constraints that prioritize sensors with high queue lengths (to prevent buffer overflow) and favor good channel conditions. Based on the task description depicted in \ref{fig:task}(d), the LLM generates power adjustment recommendations that are implemented, with the resulting system performance and last environmental state being monitored as feedback. This continuous feedback loop,  tracking the LLM's power decisions, the latest sensor data, and transmission outcomes, enables dynamic refinement of power settings, allowing the system to progressively improve energy efficiency while maintaining reliable data transmission and preventing queue overflows. LLM-assisted ICL offers a lightweight alternative to DRL for power control in UASNETs, where it replaces retraining with example-based reasoning and enables real-time decision-making with minimal computational overhead, which is especially well-suited for dynamic, resource-constrained, and safety-critical environments where DRL's complexity and retraining demands become limiting factors.
In this sensor-side power control task, the cost reflects the overall energy efficiency of the network. It is derived from the total energy spent by all ground sensors divided by the number of successfully transmitted bits, together with an additional term representing the overall packet-loss ratio. These indicators are obtained directly from the sensors’ power telemetry and communication logs and are incorporated into the feedback context for subsequent LLM reasoning.
\begin{figure*}[h!]
    \centering
    \begin{subfigure}[b]{0.3\textwidth}
        \centering
        \includegraphics[height=4cm, width=\linewidth, keepaspectratio]{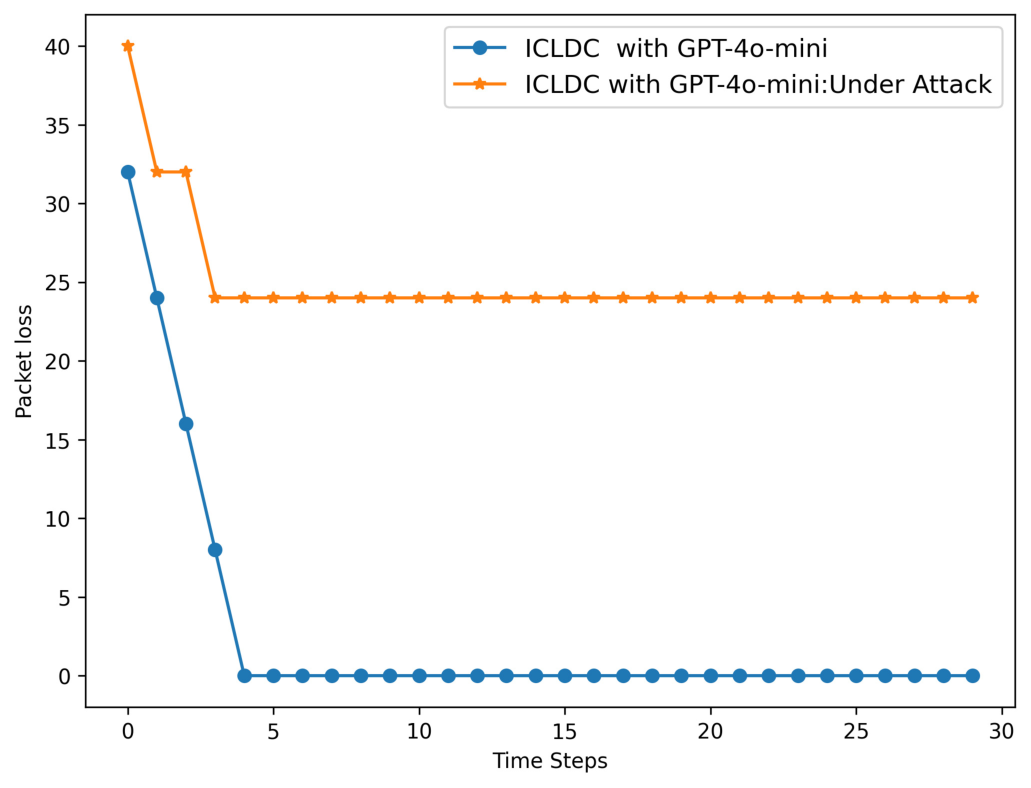}
        \caption{Normal and under-attack operation of ICLDC.}
        \label{fig:sub1}
    \end{subfigure}
    \hfill
    \begin{subfigure}[b]{0.3\textwidth}
        \centering
        \includegraphics[height=4cm, width=\linewidth, keepaspectratio]{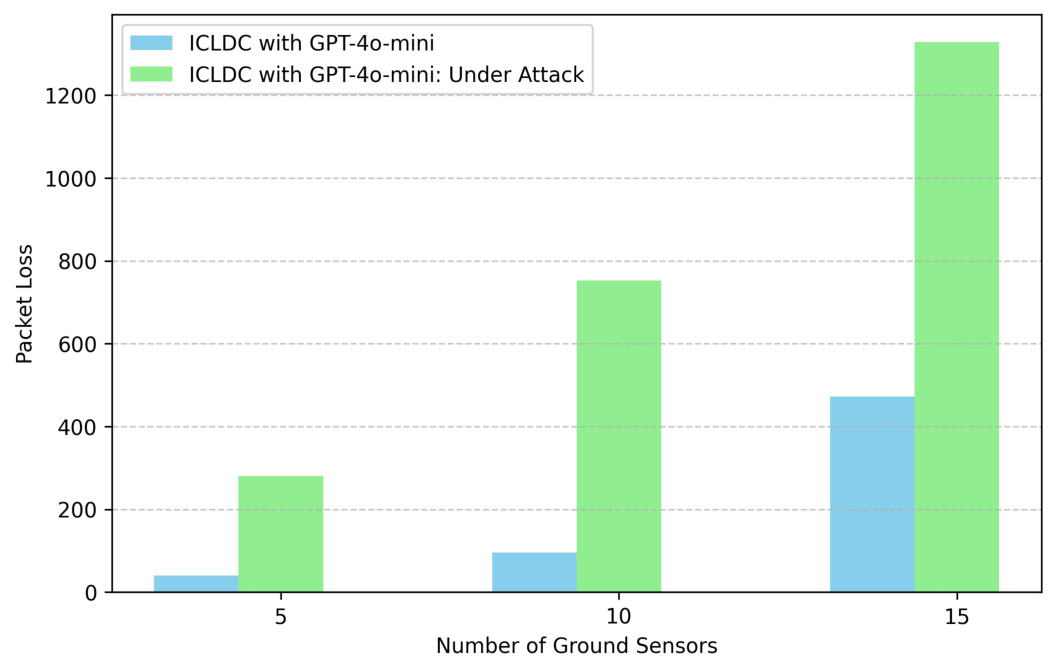}
        \caption{Performance with varying numbers of ground sensors.}
        \label{fig:sub2}
    \end{subfigure}
    \hfill
    \begin{subfigure}[b]{0.3\textwidth}
        \centering
        \includegraphics[height=4cm, width=\linewidth, keepaspectratio]{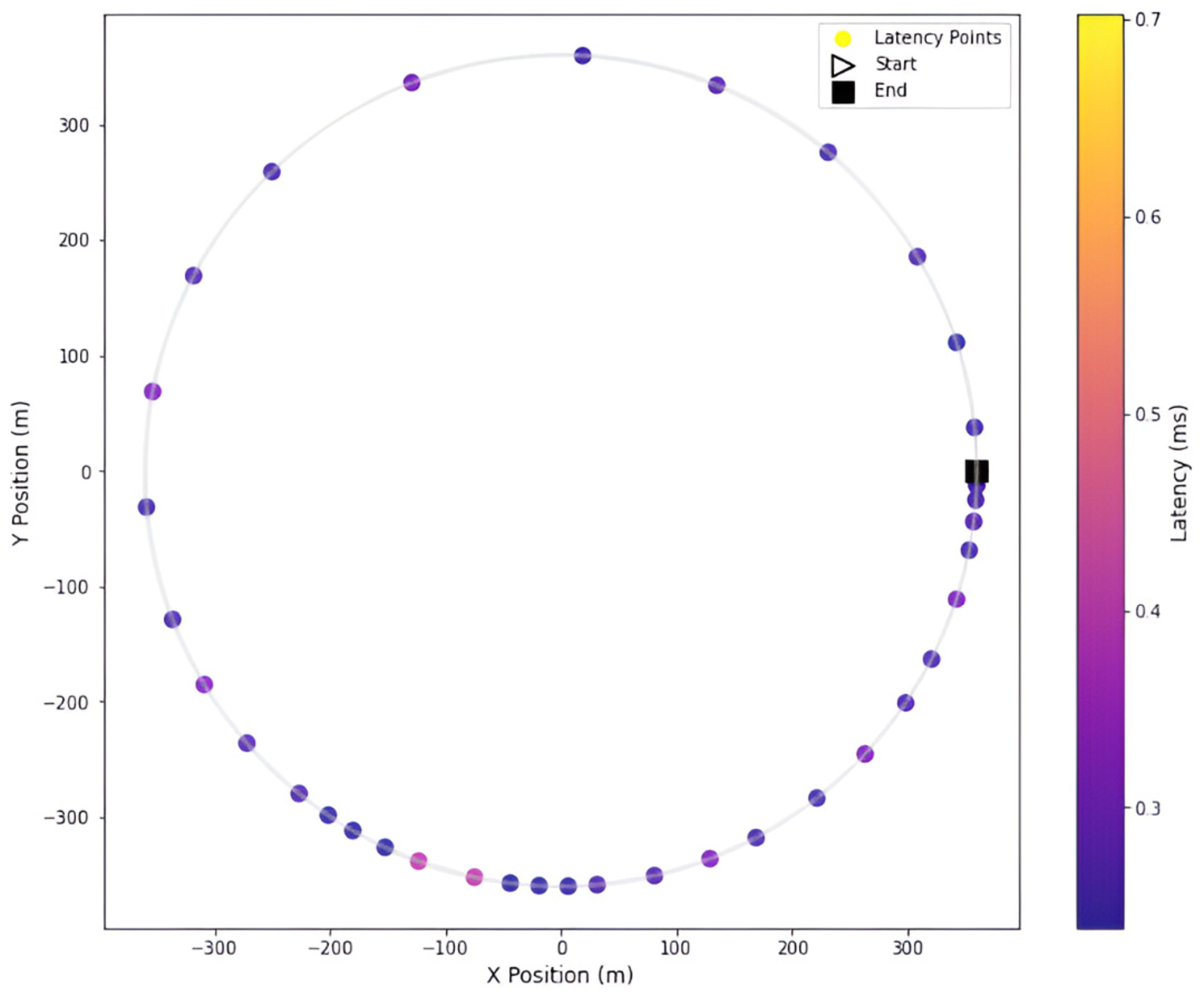}
        \caption{Measured LLM inference latency heatmap on UAV circular flight path.}
        \label{fig:sub3}
    \end{subfigure}

    \caption{Performance analysis of ICLDC in multiple scenarios with 10 ground sensors.}
    \label{fig:three_in_row}
\end{figure*}

\section{Case Study: Data Collection Schedule} \label{sec5}
This section presents a case study focused on data collection scheduling. The objective is to demonstrate how ICL effectively manages data collection scheduling and is vulnerable to jailbreaking attacks. Our case study introduces a unique security vulnerability through its reliance on contextual prompts. To address this, a formal black-box attacker model is defined, where an adversary with no internal model access manipulates the system by injecting a malicious demonstration into the prompt context.

\begin{table}[h]
\centering
\caption{Runtime and Complexity Analysis of ICLDC}
\label{tab:comparison20}
\begin{tabular}{|l|c|}
\hline
\textbf{Metric} & \textbf{ICLDC} \\ \hline
Training Episodes & 1 (30 steps) \\ \hline
Training Time & $\sim$20 seconds (CPU) \\ \hline
Sample Efficiency & 30 interactions \\ \hline
Compute Requirements & Low (CPU-only) \\ \hline
Computational Complexity & $\mathcal{O}(n^2 + k \cdot n)$ \\ \hline
\end{tabular}

\end{table}

The attacker's objective is to alter the LLM's output distribution to maximize a network-level loss function, such as one based on packet loss, thereby degrading system performance by forcing suboptimal decisions. We assume the attack is executed through methods such as jailbreaking via malicious prompts. An ICL-based Data Collection Scheduling (ICLDC) framework for UASNETs is proposed in \cite{emami2025llm}. The proposed ICLDC leverages ICL to optimize the UAV's emergency data collection schedules. Specifically, the UAV gathers sensory information, such as queue lengths and channel conditions, from ground sensors and transmits this data to an edge-hosted LLM. The LLM is initialized with a structured system prompt that outlines its operational role, constraints, and ethical safeguards, ensuring compliance with safety and security requirements. Upon receiving the environmental data, the LLM generates a detailed task description in natural language, encompassing the mission objective, input parameters, operational rules, illustrative examples, expected outputs, and a feedback mechanism.

Based on this context, the LLM infers an optimal data collection schedule, which the UAV subsequently executes. Performance metrics and updated environmental states are continuously monitored and incorporated as feedback into the evolving task description. This feedback is utilized in subsequent decision cycles, enabling the system to adapt dynamically and improve scheduling efficacy over time. 

Moreover, the paper examines the system’s vulnerability to adversarial manipulation, specifically jailbreaking attacks, where the task description is maliciously altered to degrade scheduling performance and increase network cost. This highlights the critical need for robust safeguards when deploying LLM-assisted control schemes in mission-critical applications. Fig. \ref{fig:sub1} illustrates the performance of the proposed ICLDC under attack in comparison to the normal scenario. The results clearly show that the attack substantially degrades performance, resulting in higher costs. The attacker employs context-based manipulation to alter the task description within the LLM’s prompt. This deliberate interference misleads the LLM, prompting it to choose suboptimal or potentially harmful actions, which in turn disrupts the overall system behavior. Fig. \ref{fig:sub2} illustrates the performance of ICLDC compared to the under-attack baseline as the number $N$ of ground sensors increases. In general, a larger number of sensors leads to higher costs due to an increased likelihood of buffer overflow. Notably, ICLDC consistently achieves the lowest cost, demonstrating its effectiveness in managing network dynamics.  Fig. \ref{fig:sub3} shows a visualization of the LLM (GPT-4o-mini) inference latency measurements along a circular trajectory. The latency values at each point along the trajectory are represented by a color gradient, with different hues corresponding to different latency intensities measured in milliseconds (ms). The visualization effectively shows how LLM inference latency varies at different positions along the circular trajectory.

Table \ref{tab:comparison20} shows that ICLDC requires only 1 training episode (30 steps), executes in approximately 20 seconds on a CPU, and achieves high sample efficiency (30 interactions), all while operating with low compute requirements. In addition, the computational complexity of ICLDC depends on both the length of the input sequence and the dimensionality of the model. Specifically, the cost increases with the square of the input sequence length, reflecting the overhead of processing interactions between elements in the sequence, and scales linearly with the model dimension for each input token. This characterization captures the computational cost of querying the language model at each decision step, providing a theoretical basis for evaluating the efficiency and scalability of the framework.

\begin{table*}[h]
\centering
\caption{Comparison of Optimization Approaches for Public Safety UAV}
\label{tab:comparison}
\begin{tabular}{|p{3cm}|p{4.5cm}|p{4cm}|p{4cm}|}
\hline
\textbf{Feature} & \textbf{DRL Optimizers} & \textbf{DM + DRL Hybrid} & \textbf{LLM Optimizer (ICL)} \\
\hline
\textbf{Training Requirements} & Complex model training (1,000 episodes, 30,000 steps; $\sim$50 min on GPU) & Improved Model Training & No retraining or parameter updates (1 episode, 30 steps; $\sim$20 sec on CPU)\\
\hline
\textbf{Deployment Speed} & Slow deployment & Improves real-time performance & Rapid adaptability and deployment \\
\hline
\textbf{Sample Efficiency} & Low (30,000 interactions) & Improved & High (30 interactions; 1,000$\times$ more efficient)\\
\hline
\textbf{Compute Requirements} & High (GPU needed) & Moderate & Low (CPU-only; no GPU required) \\
\hline
\textbf{Real-World Gap} & Gaps between simulation and deployment & Creates realistic simulation environments & Natural language problem expression \\
\hline
\textbf{Data Dependency} & High & Alleviates data scarcity & Problem described in natural language \\
\hline
\textbf{Adaptability} & Action-oriented, lacks flexibility & Improves adaptability & Facilitates rapid task adaptation \\
\hline
\textbf{Inference Latency} & Low & Low & 0.3–0.7 ms (average 0.3 ms)\\
\hline
\textbf{Key Advantage} & State-of-the-art for UAV-assisted data collection & Robust solution for complex scenarios & Flexible, generalizable, highly sample-efficient \\
\hline
\textbf{Key Limitation} & Slow training and deployment & Still requires complex integration & Unproven convergence properties\\
\hline
\end{tabular}

\end{table*}

\section{Discussion: From DRL to LLM Optimizers} \label{sec6}
This section discusses the limitations of DRL algorithms and the need for LLM optimizers like ICL in UASNETs for emergency scenarios. DRL algorithms are considered state-of-the-art optimizers for UASNETs. However, DRL-enabled optimizers tend to be very action-oriented and suffer from challenges such as complex model training and gaps between simulation and real-world deployment, which hinder their rapid deployment in emergencies. In particular, the time-consuming nature of model training and deployment is not compatible with the urgent requirements of public safety situations. What we need are optimizers that offer rapid adaptability and deployment to meet the critical requirements of emergency scenarios.
\par
A promising approach to complement DRL algorithms is the integration of DMs. By combining the data generation capabilities of DMs with the decision-making framework of DRL, this integration improves the adaptability and real-time performance of UASNETs. In addition, studies have shown how DMs can alleviate data scarcity, improve policy networks, and create simulation environments that provide a robust solution for complex UAV communication scenarios\cite{emami2025diffusion}. However, even with these advances, there is still a need for lighter-weight solutions that can overcome the challenges of complex model training and the gap between simulation and real-world operations. Rapid adaptability and simplicity of deployment remain key factors for UASNETs in emergencies where time and efficiency are critical.
\par
Recent advancements in prompting techniques have significantly enhanced the performance of LLMs across diverse domains. Leveraging their intrinsic ability to comprehend and generate natural language, LLMs introduce a novel paradigm for addressing optimization problems. Conventional optimization approaches typically require explicit formalization of the problem and the derivation of update rules through specialized solvers. In contrast, LLM-assisted optimization frameworks allow the problem to be expressed in natural language, enabling the model to iteratively propose candidate solutions based on the problem description and previously generated outputs.
\par
The LLM-assisted optimization framework offers several important advantages. First, it facilitates rapid adaptation to different UAV tasks by simply modifying the problem description contained in the prompt. Second, the iterative nature of the process enables continuous refinement of solutions without the need to retrain or change the underlying model parameters. Finally, the optimization process can be effectively customized through prompt engineering by including specific instructions that enforce desired solution properties such as feasibility constraints or performance criteria.

Such ICL capabilities enable LLMs to act as flexible and generalizable optimization agents. This change has the potential to accelerate the development of solutions in UASNETs where explicit mathematical modeling is a challenge. However, further research is needed to accurately quantify the convergence properties and solution quality that can be achieved by LLM-based optimization frameworks. 
Specifically, LLM-assisted optimization remains challenging due to the autoregressive nature of inference and non-stationary context distributions. While we therefore avoid theoretical claims of convergence, the proposed framework does not rely on guaranteed convergence in practice. Instead, it operates under guarded conditions: i) Externally measured and bounded performance signals, such as packet loss, queue backlog and energy, are used as feedback; ii) The high-level action space is discretized and filtered by rule-based safety constraints; iii) The efficient optimization scheme avoids the complexity of dedicated model pre-training and fine-tuning, allowing for efficient correction and control. 
Together, these mechanisms provide operational stability and safety for edge-assisted UAV control.

Developing advanced memory architectures is essential to enhance the stability and contextual consistency of LLM-assisted control during extended missions. The current framework relies on feedback-driven iterative refinement, but its contextual understanding is limited to near-term interactions. For prolonged operations, maintaining coherent reasoning and avoiding repetitive strategies requires a structured representation of mission history.

This can be addressed through a Conversation Memory Module comprising two complementary components: i) Short-Term Working Memory: A dynamic buffer that retains recent sequences of actions and outcomes to ensure continuity and coherence in decision-making; ii) Long-Term Summary Memory: A condensed knowledge base that captures and abstracts essential patterns, lessons, and strategic insights accumulated throughout the mission.
By integrating these components, the system can perform context-aware prompt retrieval, leveraging both immediate context and historical understanding. This approach supports sustained strategic awareness and reasoning stability over extended deployments, improving reliability and adaptability in complex public safety operations.
Overall, while DRL algorithms are state of the art for optimizing UASNETs, they reach their limits in emergency scenarios due to their slow training, complex application and gaps between simulation and reality, making them unsuitable for time-critical situations. DMs improve DRL by generating synthetic data and improving the adaptability of policies, but are still computationally intensive. In contrast, LLM optimizers that use ICL offer a promising alternative. They enable fast, training-free adaptation through natural language prompts, iterative refinement of the solution, and optimization that can be interpreted by humans - important advantages for emergency response. However, LLM-based approaches require further validation of their convergence properties and real-world reliability, suggesting a potential for hybrid systems that combine the robustness of DRL with the flexibility of LLM to meet urgent public safety needs. Table \ref{tab:comparison} Compare Optimization Approaches for Public Safety UAV and provide details on training time and inference latency.

\section{Most recent trends} \label{sec7}
This section highlights the future direction of LLM-assisted ICL and the ways to improve the performance and deployment.
\par
\subsubsection{\textbf{Enriching Examples using Diffusion Models}}
Focus on enhancing the generalization capabilities of public safety UAV by addressing the scarcity of high-quality training data. Develop methods to generate realistic synthetic examples using DMs to augment limited emergency datasets. Further research should explore optimizing DM-based data generation to ensure the quality of synthetic examples for robust generalization.

\subsubsection{\textbf{On-device LLMs for Public Safety UAV}}
Advance the deployment of LLMs on resource-constrained edge devices, such as UAV platforms, by leveraging recent progress in model compression, efficient architectures, and hardware-software co-optimization. Focus on enabling real-time, localized LLM processing on UAVs to eliminate reliance on remote cloud servers. This will facilitate the creation of personalized, context-aware, and low-latency AI functionalities in UASNETs. 

\subsubsection{\textbf{Enhanced Robustness Against Adversarial Attacks}}
Investigate and develop advanced defense mechanisms to mitigate vulnerabilities such as jailbreaking attacks, which manipulate prompts to bypass built-in safety mechanisms and cause LLMs to generate malicious content. Address both white-box and black-box categories of adversarial attacks to ensure the reliability of LLM-assisted ICL in mission-critical scenarios.

\subsubsection{\textbf{Extension to Public Safety UAV Swarms}}
Extend the application of LLM-assisted ICL to public safety UAV swarms, focusing on facilitating coordination among multiple UAVs. 
Develop collaborative reasoning strategies that allow UAVs to share contextual information and jointly adapt to dynamic environments. 
Develop methods that enable cooperative task execution in the swarm, leveraging the contextual understanding and adaptive reasoning capabilities of LLMs to improve overall swarm performance and mission success. 
Such an extension can support distributed decision-making, enhance resilience to single-node failure, and enable collective intelligence across heterogeneous UAV platforms.

\subsubsection{\textbf{Real-World Validation and Transferability}}
Conduct extensive field trials to evaluate the performance of LLM-assisted ICL in real-world public safety applications. 
These trials should assess latency, energy efficiency, communication reliability, and overall mission success under practical operating constraints. 
Explore the transferability of LLM-assisted ICL to other UAV applications, such as agriculture and infrastructure inspection, by repurposing task descriptions and prompts. 
Our future work will also focus on quantifying how well the proposed framework generalizes across domains and identifying the degree of prompt adaptation required for new operational contexts.

\section{Conclusion} \label{sec8}
This paper advocates integrating LLM-assisted ICL into public safety UAV to address key operational functions, such as path planning and velocity control. By leveraging ICL, UAVs can dynamically interpret task objectives, adjust behaviors, and make context-aware decisions based on natural language instructions and a few example prompts without the need for retraining. By classifying critical UAV functions and formulating corresponding task descriptions, we demonstrate how LLMs can effectively support path planning, velocity control, data collection scheduling, and power control. The proposed LLM-assisted ICL framework offers a lightweight, adaptable, and scalable approach, setting the stage for new generation of intelligent, edge-enabled public safety UAVs. We envision the application of ICL to public safety UAV swarms, along with its real-world validation and potential use in other domains.

\bibliographystyle{IEEEtran}
\bibliography{references}

\end{document}